\documentclass[journal]{IEEEtran}

\usepackage[english]{babel}
\usepackage{cite}
\usepackage{lipsum}
\usepackage{amsmath,amssymb,amsfonts}
\usepackage{graphicx}
\usepackage{textcomp}
\usepackage{xcolor}
\usepackage{float}
\usepackage{subcaption}
\usepackage{algorithm}
\usepackage{algpseudocode}
\usepackage{mwe}
\usepackage[flushleft]{threeparttable}
\usepackage{color}
\usepackage{colortbl}
\usepackage{booktabs}
\usepackage{multirow}
\usepackage{url}
\usepackage{xcolor}
\newcommand{\ra}[1]{\renewcommand{\arraystretch}{#1}}



\def\BibTeX{{\rm B\kern-.05em{\sc i\kern-.025em b}\kern-.08em
    T\kern-.1667em\lower.7ex\hbox{E}\kern-.125emX}}
    
\ifCLASSINFOpdf
\else
\fi
\begin{document}
% \linenumbers
% \linenumbersep 3pt\relax
% LIMIT THE NUMBER OF AUTHORS IN CITATIONS (and other changes, see the .bib file for this entry)
\bstctlcite{IEEEexample:BSTcontrol}

\title{Deep Learning Training on the Edge with Low-Precision Posits}

\author{Hamed F. Langroudi,~%\IEEEmembership{Student Member,~IEEE,}
        Zachariah Carmichael,~%\IEEEmembership{Student Member,~IEEE,}
        Dhireesha Kudithipudi%\IEEEmembership{Senior Member,~IEEE}% <-this % stops a space
\thanks{Hamed. F. Langroudi, Zachariah Carmichael,  and Dhireesha Kudithipudi are with the Nueromorphic AI Lab, Department
of Computer Engineering, Rochester Institute of Technology, Rochester,
NY, USA}}

% ~Vol.~TODO, No.~TODO, 
\markboth{}%
{Langroudi \MakeLowercase{\textit{et al.}}}


\IEEEspecialpapernotice{(Preprint)}

\maketitle

\begin{abstract}
%Despite this proficient performance, deploying DNNs for training and inference on edge devices, such as mobile or IoT devices, is an arduous task due to the trade-off between accuracy and energy consumption on edge devices. Accordingly, 
%To realize deep neural networks (DNNs) training and inference on edge devices, 
%Low-precision DNNs have been extensively explored with the goal of condensing DNN models for edge devices. 
Recently, the posit numerical format has shown promise for DNN data representation and compute with ultra-low precision ($[5..8]$-bit). However, majority of studies focus only on DNN inference. In this work, we propose DNN training using posits and compare with the floating point training. We evaluate on both MNIST and Fashion MNIST corpuses, where 16-bit posits outperform 16-bit floating point for end-to-end DNN training.
%Results indicate that DNN training with 32-bit and performing inference with 5-bit posit improves the energy-delay-product by $\sim$62.5\% over 32-bit float training and inference, for similar accuracy.
\end{abstract}

% Note that keywords are not normally used for peer review papers.

\IEEEpeerreviewmaketitle

\section{Introduction}
%Deep learning has propelled forwards the use of machine learning/artificial intelligence in a vast spectrum of applications in various domains, such as computer vision \cite{DensNetCVPR}, natural language processing \cite{GNMT}, speech recognition \cite{DeepSpeech3}, healthcare \cite{hannun2019cardiologist}, agriculture \cite{olsen2019deepweeds}, transportation \cite{dabiri2019semi}, and cybersecurity \cite{sirinam2018deep}. 
%[D will rewrite this section] Currently, most machine learning workloads in these applications are performed in a centralized computation model, such as data center infrastructure, because of the demand for processing massive amounts of data rapidly with minimum latency. 
%However, in many applications the centralized computation model is not optimal due to privacy issue caused by transferring data to/from the data center, unreliable connection between edge devices and data center imposed by the environment where the application is located, and system failure issues by missing real-time constraints where the data transferring latency is bigger than the data analysis latency \cite{Edge,xu2018scaling}. For instance, constructing a reliable connection to transfer data from edge device to the data center to perform deep learning on agriculture application is a rigorous task. Therefore, decentralized and distributed computation models, such as Edge computing, are suitable for these types of applications. 

The edge computing, offers a decentralized solution to cloud-based datacenters \cite{2018edge} and intelligence-at-the-edge of mobile networks. However, training on the edge is a challenge for many deep neural networks (DNNs). This arises due to the significant cost of multiply-and-accumulate (MAC) units, an ubiquitous operation in all DNNs. In a 45~nm CMOS process, energy consumption doubles from 16-bit floats to 32-bit floats for addition and by ${\sim}4\times$ for multiplication \cite{horowitz20141}. Memory access cost increases by ${\sim}10\times$ from 8~kB to 1~MB memory with a 64-bit cache \cite{horowitz20141}. In general, there is a gap between memory storage, bandwidth, compute requirements, and energy consumption of modern DNNs and hardware resources available on edge devices \cite{facebookEdge}.

An apparent solution to address this gap is to compress such networks, thus reducing the compute requirements to match putative edge resources. Several groups have proposed compressed new compute- and memory-efficient DNN architectures \cite{MobileNet,chen2019drop,MEC} and parameter-efficient neural networks, using methods such as DNN pruning \cite{ren2018sbnet}, distillation \cite{zhou2018revisiting}, and low-precision arithmetic \cite{Jacob_2018_CVPR,IBM8}. Among these approaches, low-precision arithmetic is noted for its ability to reduce memory capacity, bandwidth, latency, and energy consumption associated with MAC units in DNNs and an increase in the level of data parallelism \cite{Jacob_2018_CVPR, hashemi2017, gysel2018}.

%For example, in case of visual recognition tasks performed on mobile devices: if there are  12,000 users transmitting 1080p video, then one will require a link of 100 gigabits/s. If it scales to a million users then that would translate to 8.5 terabits/s \cite{satyanarayanan2017emergence}, which becomes untenable when performing complex tasks. Similarly, a modern self-driving vehicle generates anywhere between 3 GB/s - 40GB/s from the different sensors for rapid analytics \cite{flashmemorysummit,maughan2019no} and a modern aircraft generates $\sim${0.5 TB} data during a \textcolor{red}{flight} \cite{satyanarayanan2017emergence}. 

%It is therefore increasingly critical to design systems that can efficiently perform intelligence-on-the-edge with minimal resources at scale.

%To address this, several groups have preserved accuracy with new compute-and memory-efficient neural networks \cite{MobileNet,chen2019drop,MEC} and parameter-efficient neural networks, such as DNN pruning \cite{ren2018sbnet}, distillation \cite{zhou2018revisiting}, and low-precision arithmetic \cite{Jacob_2018_CVPR,IBM8}. 

%Among the techniques to compress neural network parameters, low-precision arithmetic is noted for its ability to reduce memory capacity, bandwidth, latency, and energy consumption associated with MAC units, and increase the level of data parallelism \cite{Jacob_2018_CVPR, hashemi2017, gysel2018}. 

The ultimate goal of low-precision DNN design is to reduce the original hardware complexity of the high-precision DNN model to a level suitable for edge devices without significantly degrading performance.

%TODO: Add paragraph
To address the gaps in previous studies, we are motivated to study low-precision posit for DNN training on the edge.

% \section{Background}
\section{Posit Numerical Format}

% \subsection{Posit Numerical Format}

An alternative to IEEE-754 floating point numbers, posits were recently introduced and exhibit a tapered-precision characteristic \cite{gustafson2017beating}. Posits, a Type III unum, offer an elegant resolution to many of the shortcomings of IEEE-754 floating format and address limitations of both Type I and Type II unums \cite{tichy2016unums}. Moreover, posits provide better accuracy, dynamic range, and program reproducibility than IEEE floating point. The essential advantage of posits is their capability to represent nonlinearly distributed numbers in a specific dynamic range about unity ({$\pm$}1.0) with high accuracy.
The value of a posit bit-string is governed by \eqref{equ:equ00}, where $s$ represents the sign, $es$ the maximal exponential bits, $fs$ the maximal fractional bits, $e$ the exponent value, $f$ the fraction value, and $k$ the regime value (as given by \eqref{equ:equ01}).
\begin{equation}
    x= 
\begin{cases}
    0, & \text{if } ({\tt 00...0})   \\
    NaR, & \text{if }  ({\tt 10...0}) \\
    (-1)^{s}\times 2^{2^{es} \times k} \times 2^e \times \left(1+\frac{f}{2^{fs}} \right), & \text{otherwise}
\end{cases}
\label{equ:equ00}
\end{equation}
The regime is encoded based on the \emph{runlength} $m$ of identical bits $(r...r)$ terminated by either a \emph{regime terminating bit}~$\overline{r}$ or the end of the bit-string of size $n$. Note that there is no requirement to distinguish between negative and positive zero since only a single bit pattern ${\tt (00...0)}$ represents zero. Furthermore, instead of defining a ``Not-a-Number'' ($NaN$) for exceptional values and infinity by various bit patterns, a single bit pattern $({\tt 10...0})$, ``Not-a-Real'' ($NaR$), represents all such values. Furthermore, $NaR$ never arises due to overflow or underflow. More details about the posit numerical format can be found in \cite{gustafson2017beating}.
\begin{equation}
    k= 
\begin{cases}
    -m , & \text{if } r= 0\\
     m-1, & \text{if } r= 1
\end{cases}
\label{equ:equ01}
\end{equation}

\section{Related Work}

%As lately as 1980s, low-precision arithmetic has been studied for shallow neural networks to not only reduce compute and memory complexity on performing training and inference without scarifying the performance, but in some scenarios also improves the performance of training and inference when the noise generated by the low-precision of shallow neural network parameters acts as regularization \cite{iwata1989artificial,hammerstrom1990vlsi}.
%The outcomes of these studies indicate that 8- to 16-bit precision is sufficient for DNN training and inference on shallow networks \cite{iwata1989artificial,hammerstrom1990vlsi}. The capability of using low-precision arithmetic to reduce memory footprint, and energy consumption of performing DNN training and inference are reevaluated in the deep learning era \cite{Gupta,Courbariaux14}. Several of these studies have shown that 16-bit/32-bit mixed-precision floating point and 16-bit fixed-point are sufficient to perform DNN training and inference without retraining respectively within the same accuracy of 32-bit floating point \cite{judd2016proteus,pudiannao2015}. 

As early as the 1980s, low-precision arithmetic has been explored in shallow neural networks to decrease both compute and memory complexity for training and inference without deteriorating performance \cite{graf1988vlsi,iwata1989artificial,hammerstrom1990vlsi, asanovic1991experimental}. In some scenarios, this bit-precision constraint also improves DNN performance due to the quantization noise acting as a regularization method \cite{asanovic1991experimental,bishop1995training}.
The outcome of these studies indicate that 16- and 8-bit precision DNN parameters are capable of satisfactorily maintaining performance for both training and inference in shallow networks \cite{iwata1989artificial,hammerstrom1990vlsi,asanovic1991experimental}. The capability of low-precision arithmetic is reevaluated in the deep learning era to reduce memory footprint and energy consumption during training and inference \cite{Courbariaux14,Gupta,micikevicius2017mixed,flexpoint2017,IBM8,mellempudi2019mixed,kalamkar2019study,gysel2018,hashemi2017,Microsoft2018,carmichael2019positron,carmichael2019performance,Hamed2018,johnson2018rethinking}. 

In performing DNN training, several previous studies utilize either variants of low-precision block floating point (BFP) (blocks of floating point DNN parameters that share an exponent \cite{wilkinson1965rounding})
% such as Flexpoint \cite{flexpoint2017},
% (16-bit fraction with 5-bit shared exponent for DNN parameters),
or mixed-precision floating point.
% (16-bit weights, activations, and gradients and 32-bit accumulators in the SGD weight update process)
These methods are sufficient to maintain similar performance as 32-bit high-precision floating point. For instance, Courbariaux \textit{et al.} trained a low-precision DNN on the MNIST, CIFAR-10, and SVHN datasets with the floating point, fixed-point, and BFP numerical formats \cite{Courbariaux14} and show that BFP achieves the best performance due to variability between the dynamic range and precision of DNN parameters \cite{Courbariaux14}. Following, Koster \textit{et al.} proposed the Flexpoint numerical format and a new algorithm, Autoflex, which analyzes the statistics of the history of DNN parameters, to optimally select the shared exponents for DNN parameters iteration-wise during gradient descent \cite{flexpoint2017}.

Aside from
% managing the shared exponent in
the BFP numerical format, Narang \textit{et al.} explored mixed-precision floating point \cite{micikevicius2017mixed} using 16-bit floating point weights, activations, and gradients during both the forward and backward passes. To prevent accuracy loss caused by underflow
% in the product of learning rate and gradients 
% with \eqref{eq:GradientDescent}
in 16-bit floating point, the weights are updated with 32-bit floating point. Additionally, to prevent gradients with very small magnitude from becoming zero when represented by 16-bit floating point, a new loss scaling approach is proposed.

Recently, Wang \textit{et al.} and Mellempudi \textit{et al.} propose a method to reduce the bit-precision of weights, activations, and gradients to 8 bits by exhaustively analyzing DNN parameters during training \cite{IBM8,mellempudi2019mixed}. In \cite{mellempudi2019mixed}, a new chunk-based addition is presented to solve the truncation issue caused by the addition of large- and small-magnitude numbers, thus successfully reducing the number of bits for the accumulator and weight updates to 16 bits. To mitigate requiring loss scaling in mixed-precision floating point training, Kalamkar \textit{et al.} \cite{kalamkar2019study} proposed the brain floating point (BFLOAT-16) half-precision format with a reduced 8-bit fractional precision
and similar dynamic range (7-bit exponent)
to 32-bit floating point. A side effect of this representation is that the conversion complexity between these BFLOAT-16 and IEEE floating point is reduced during training. In training a ResNet model on the ImageNet dataset, BFLOAT-16s achieve the same performance as 32-bit floating point.

This research builds on earlier studies \cite{carmichael2019positron,carmichael2019performance,Hamed2018,johnson2018rethinking,montero2019template} and for the first time studies feedforward neural network training with posits on MNIST and Fashion MNIST datasets .

%This paper is inspired by the earlier studies and tries to extend \cite{carmichael2019positron} for convolutional deep neural networks and higher-dimensional datasets, such as MNIST and CIFAR-10, and explores the efficacy of posit numerical format for especially $\leq$8-bit as a extension of \cite{johnson2018rethinking}.

% TODO : needs to rewrite  
\section{Proposed Work, Results \& Analysis}
\begin{figure*}%[h!tb]
\centering
\includegraphics[width=0.6\linewidth]{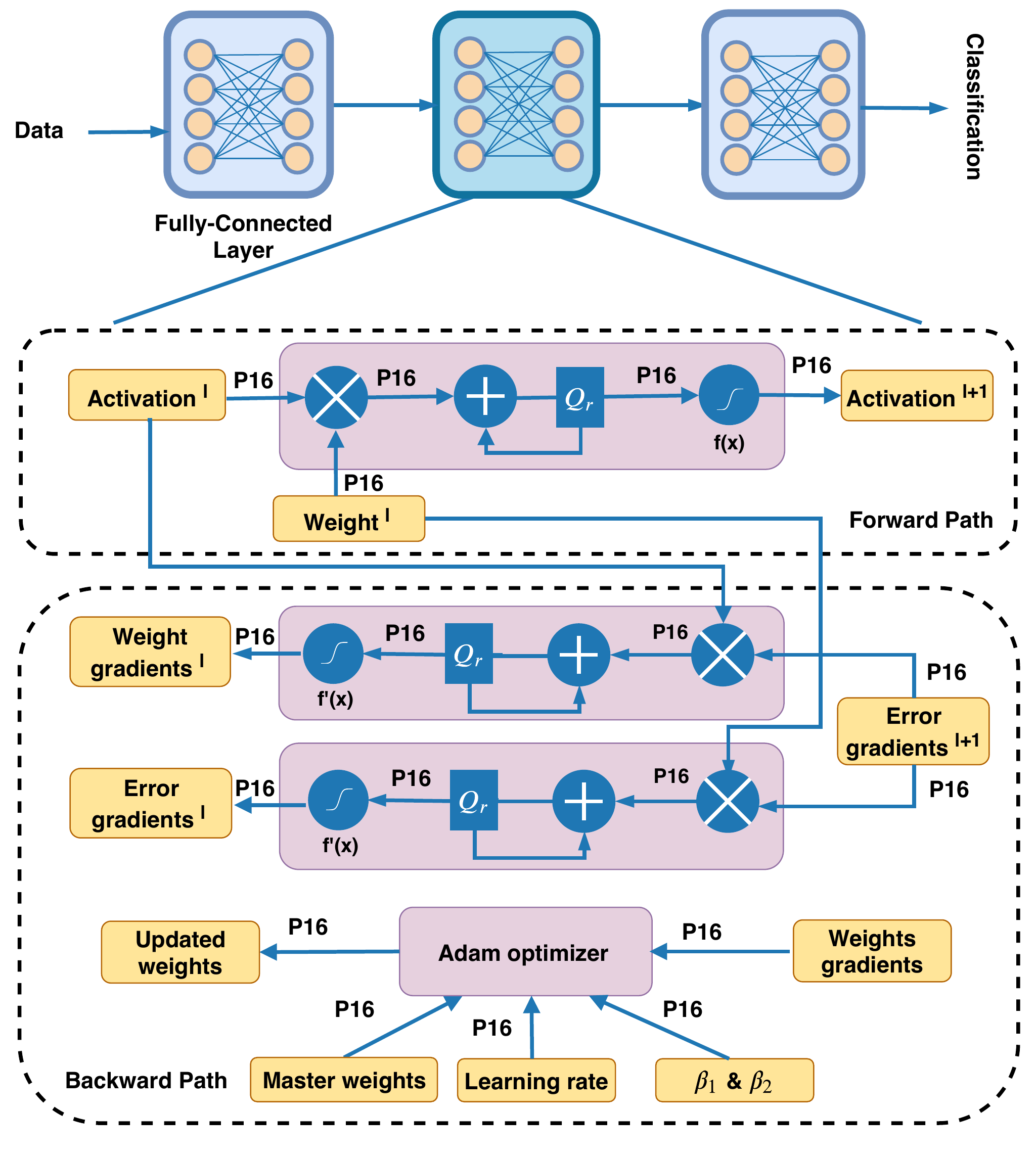}
\caption{The software framework for training the feedforward neural networks with three hidden layers using posit numerical format. The framework scales to any DNN architecture. P16: Posit16; $\beta_1 \& \beta_2$: Exponential decay rates. }
\label{fig:Software-Cheetah}
\vspace{-3mm}
\end{figure*}

To perform DNN training in feedforward neural networks, we study two numerical formats (floating point and posit) with 32-bit and 16-bit precision. For simplicity, the
% \emph{Cheetah} % <-- removed
architecture % <-- added
explained here is based on three hidden layers feedforward neural network training with the posit numerical format as shown in Fig. \ref{fig:Software-Cheetah}.

% this is the first version
% s/w framework and h/w framework is confusing.Software specification of system? 
% \subsection{Software Design and Exploration}
% % To emulate the feedforward neural network, each layer's output is calculated as
%In emulating feedforward networks, the output of each layer $Y$ is calculated in the forward path as given by \eqref{eq:FeedForwardNetwork-Convolution-Quantized}
% \begin{equation}
% Y_j{=}B_j{+}\frac{1}{\alpha_1{\times}\alpha_2}{\times}
% \left(
%     \sum_{i}^{N}
%     \left[ Q(\alpha_1{\times}A_i) \right]
%     {\times}
%     \left[ Q(\alpha_2{\times}W_{ij}) \right]
% \right)
% \label{eq:FeedForwardNetwork-Convolution-Quantized}
% \end{equation}
%\begin{equation}
%Y_j =Q(B_j){+}
%    \sum_{i}^{N}
%    Q(A_i)
%    {\times}
%    Q(W_{ij})
%label{eq:FeedForwardNetwork-Convolution-Quantized%}
%\end{equation}

\noindent
The networks are implemented in the Keras \cite{keras} and TensorFlow \cite{tensorflow2015} frameworks. \{16, 32\}-bit floating point and posit numbers for DNN training are extended to these frameworks via software emulation.

To compare the posit and floating point numerical formats, a four-layer feedforward neural network is trained with each of the number system on the MNIST and Fashion-MNIST datasets. The results indicate that posits have improved accuracy in comparison to floating point at both 16- and 32-bit precision, as shown in Table \ref{tab:train_acc}.
% 16-bit posits outperform 16-bit floats in terms of accuracy.
Although the evaluation is demonstrated on small datasets, there are two advantages compared to \cite{IBM8,mellempudi2019mixed}. Mellempudi \textit{et al.} \cite{mellempudi2019mixed} use 32-bit numbers for accumulation to reduce the hardware cost of stochastic rounding. Wang \textit{et al.} \cite{IBM8} reduce the accumulation bit-precision to 16 by using stochastic rounding. However, in this paper, we show the potential of using 16-bit posits for \textit{all DNN parameters} with a simple and hardware-friendly round-to-nearest algorithm and show less than 2\% accuracy degradation without exhaustively analyzing the network training parameters.

\begin{table}[H]
\caption{Average accuracy over 10 independent runs {on the test set of the respective dataset. Networks are trained using only the specified numerical format}.}\label{tab:train_acc}
\centering
\ra{1.2}
\begin{threeparttable}
% \resizebox{.85\linewidth}{!}{%
\begin{tabular}{@{}lcc@{}}
    \toprule
    Task          & Format   & Accuracy \\
    \midrule
    \multirow{4}{*}{MNIST}         & Posit-32 & \textbf{98.131\%} \\
                                   & Float-32 &         98.087\%  \\
                                   & Posit-16 &         96.535\%  \\
                                   & Float-16 &         90.646\%  \\[+1ex]
    \multirow{4}{*}{Fashion MNIST} & Posit-32 & \textbf{89.263\%} \\
                                   & Float-32 &         89.105\%  \\
                                   & Posit-16 &         87.400\%  \\
                                   & Float-16 &         81.725\%  \\
    \bottomrule
\end{tabular}
% }
% \begin{tablenotes}
% \end{tablenotes}
\end{threeparttable}
\end{table}

A summary of recent studies that propose low-precision training frameworks are shown in Table \ref{tab:compare}. Several research groups have explored the efficacy of floats and BFP on the performance of DNNs with multiple image classification tasks \cite{micikevicius2017mixed,flexpoint2017,mellempudi2019mixed,IBM8,kalamkar2019study,montero2019template}. However, the majority of these works analyze the appropriateness of the posit numerical format for DNN training.
% Additionally, current works do not offer insight on the impact of the quantization approach vs. numerical format on both accuracy and hardware complexity, as investigated in this paper.

\begin{table*}
   \caption{High-level summary of proposed work and other low-precision training frameworks. All datasets are image classification tasks. FMNIST: Fashion MNIST; FP: floating point; FX: fixed-point; PS: posit.}
   \label{tab:compare}
   \ra{1.3}
   \centering
   \resizebox{\linewidth}{!}{%
   \begin{tabular}{@{}cccccccc@{}}
     \toprule
     & Montero \textit{et al.} \cite{montero2019template}& Narang \textit{et al.} \cite{micikevicius2017mixed} & Koster \textit{et al.} \cite{flexpoint2017} & Mellempudi \textit{et al.} \cite{mellempudi2019mixed} & Wang  \textit{et al.} \cite{IBM8} & Kalamkar \textit{et al.} \cite{kalamkar2019study} & This Work \\
     \midrule
    %  \hline
    %  Task & Image Classification & Image Classification & Image Classification & Image Classification & Image Classification & Image Classification & Image Classification \\
    %  \hline
     \multirow{2}{*}{Dataset} & \multirow{2}{*}{Synthetic} & \multirow{2}{*}{ImageNet} & \multirow{2}{*}{{CIFAR}-10} & \multirow{2}{*}{ImageNet} & \multirow{2}{*}{ImageNet} & \multirow{2}{*}{ImageNet} & \multirow{2}{*}{MNIST, FMNIST} \\[-1ex]
     &  && & &&&  \\
     \multirow{2}{*}{Numerical Format} & \multirow{2}{*}{PS} & \multirow{2}{*}{FP} & \multirow{2}{*}{BFP} & \multirow{2}{*}{FP} & \multirow{2}{*}{FP} & \multirow{2}{*}{BFLOAT} & \multirow{2}{*} {PS, FP} \\[-1ex]
     &  &  & & & & &  \\
    %  \hline
     Bit-precision & [8..16] & 16/32 & 16+5 & 8/32 & 8/16 & 16 & 16 \\ 
    %  \hline
     \multirow{2}{*} {DNN library} & \multirow{2}{*} {PySigmoid} & \multirow{2}{*} {Caffe/PyTorch} & \multirow{2}{*} {Neon} & \multirow{2}{*} {TensorFlow} & \multirow{2}{*} {Home {S}uite} & \multirow{1}{*} {IntelCaffe/Caffe2} & \multirow{2}{*} {Keras/TensorFlow} \\[-1ex]
     & & & & & & Neon/Tensorflow& \\
     \bottomrule
   \end{tabular}
   }
   \label{tab:OthersWork}
 \end{table*}

\section{Conclusions}
This work presented low-precision posit designs for both training and inference on edge devices.
% We explored the \textcolor{red}{capacity} of various numerical formats\textcolor{red}{,} including floating point, fixed-point and posit\textcolor{red}{, for} both DNN training and inference.
We show that the novel posit numerical format has high efficacy for DNN training at \{16, 32\}-bit precision training, surpassing the equal-bandwidth fixed-point and floating point counterparts.
% Moreover, we show that it is possible to achieve better performance and reduce energy consumption by using linear quantization \textcolor{red}{with} the posit numerical format.
% The success of low-precision posit\textcolor{red}{s in} reduc\textcolor{red}{ing} DNN hardware complexity with negligible accuracy degradatio\textcolor{red}{n} motivates us to evaluate ultra\textcolor{red}{-}low precision training \textcolor{red}{in} future work.
The success of posits in these experiments, needs further exploration of ultra-low precision posit training for richer datasets.

\ifCLASSOPTIONcaptionsoff
  \newpage
\fi

\bibliographystyle{IEEEtran}
\bibliography{Journal.bib}

%\end{IEEEbiography}


%\newpage



%\vfill



\end{document}